# DGAFF: Deep Genetic Algorithm Fitness Formation for EEG Bio-Signal Channel Selection

Ghazaleh Ghorbanzadeh[1], Zahra Nabizadeh[1], Nader Karimi[1], Pejman Khadivi[2], Ali Emami[1], Shadrokh Samavi[1,2,3]

[1]Isfahan University of Technology, Isfahan, 84156-83111 Iran,
[2]Computer Science Department, Seattle University, Seattle, 98122 USA
[3]Elect. & Comp. Engineering, McMaster University, L8S 4L8, Canada

**Abstract**

Brain-computer interface systems aim to facilitate human-computer interactions in a great deal by direct translation of brain signals for computers. Recently, using many electrodes has caused better performance in these systems. However, increasing the number of recorded electrodes leads to additional time, hardware, and computational costs besides undesired complications of the recording process. Channel selection has been utilized to decrease data dimension and eliminate irrelevant channels while reducing the noise effects. Furthermore, the technique lowers the time and computational costs in real-time applications. We present a channel selection method, which combines a sequential search method with a genetic algorithm called Deep GA Fitness Formation (DGAFF). The proposed method accelerates the convergence of the genetic algorithm and increases the system's performance. The system evaluation is based on a lightweight deep neural network that automates the whole model training process. The proposed method outperforms other channel selection methods in classifying motor imagery on the utilized dataset.

## 1. Introduction

Brain-Computer Interface (BCI) systems provide an external interface between the brain of a human and a digital assistive device. Such systems translate a presentation of brain cells' communication signals into digital commands [1]. In the first step, a representation of brain cells' activity is recorded as the raw data. Then, the user intention is estimated by processing the brain signals, and a relevant command is issued corresponding to the system application. In some designs, a closed-loop branch gets feedback from the performed action related to the output command and improves the system performance. Such systems assist people who suffer from movement disabilities and make their life significantly easier [2].

Various techniques may be used to acquire a presentation of brain activity. Among different technologies for recording brain signals, electroencephalography (EEG) is the most widely used method, as it is a portable, non-invasive and cost-efficient technology with high temporal resolution [3]. Recording of EEG signals in various BCI scenarios can be performed in different paradigms. Motor imagery is a category of BCI systems in which the subject should imagine the movement of body parts, and the system distinguishes which part of the body is imagined. Hence, the motor imagery BCI (MI-BCI) systems classify recorded signals into the imagination of body part movements. In some MI-BCI systems, classes are limited to four body parts: left hand, right hand, feet, and tongue [4]. Binary classification of the right and left hand, or right hand and foot is also widespread, even in the datasets with more than two classes, due to the simplicity of binary classification [5, 6].

Despite EEG benefits, this kind of brain activity recording has a poor spatial resolution [3]. Nowadays, EEG recording devices are provided with a large number of channels to augment the spatial resolution of brain activity information. But, the methodology of recording EEG signals is itself the source of the lack of high-resolution spatial information. Thus, recording the high number of channels can lead to redundancy in data [3, 7]. The redundancy in data is also a source of increasing difficulty in training a classifier. On the other hand, previous studies have shown that, although each part of the brain is involved in special tasks, there are variations between different individuals. This is another reason to select channels related to specific tasks in an inter-subject manner [7].

With the increasing use of deep neural networks in various applications, deep learning-based classifiers have become more common in BCI studies. Utilizing deep neural networks reduces the need for expert supervision by integrating feature extraction and classification. Although ideally, the network itself is expected to extract the most informative channels based on the task under the test, due to the difficulty of the data recording process and the lack of large training datasets, channel selection can be used to reduce the input dimensions. Therefore, the network is boosted to achieve higher accuracy.

Accordingly, this paper presents a channel selection method based on deep learning classifiers. In the proposed method, a proposed version of the genetic algorithm, called Deep GA Fitness Formation (or DGAFF for short), is used as the search algorithm to select the best channel subset from the possible channel subsets. The utilized fitness function uses network accuracy as the evaluation of each subset of channels. This method also uses a hierarchical search algorithm to improve the selection of the initial population of the genetic algorithm. As a result, DGAFF convergence is accelerated by the appropriate selection of the initial population.

The highlights of the paper are summarized as follows:
- We develop a channel selection method using deep learning in the Motor-Imagery (MI) classification framework.
- We propose to use a combination of the genetic algorithm and a sequential search algorithm to search among possible channel subsets for faster convergence of the genetic algorithm.
- We utilize a deep learning architecture as the proposed genetic algorithm's fitness function.
- This research has shown that the proposed framework can facilitate the training of deep-learning-based classifiers as it's a challenging task.
- Using deep learning for classification has boosted the classifier's extraction of valuable features; hence, no need for handcrafted features.
- The proposed channel selection method eliminates irrelevant channels and simplifies the BCI implementation.

The structure of the paper is as follows. In Section 2, the literature review of this study is presented. In Section 3, different parts of the proposed BCI system are explained, and the proposed channel selection is described. Section 4 provides comparative results of the proposed method and explores the effect of system parameters. Finally, in Section 5, concluding remarks are presented.

## 2. Literature Review

EEG signals have a low Signal-to-Noise Ratio (SNR). As a result, various types of noise, artifact, and distortion can influence system performance [8]. Hence, having an accurate classification in an MI-BCI system requires proper signal cleaning and preprocessing. Filtering data in the frequency domain [6, 9, 10], windowing in the time domain [6, 9], and blind source separation methods such as PCA, ICA, and MUSE [10, 11] are some of the methods proposed for signal cleaning. In addition, feature extraction, feature selection, and channel selection are other possible approaches to reduce noise and dimension of data to eliminate irrelevant features and utilize more informative ones [7, 12]. Finally, a classifier is used to classify the extracted features into target classes. Based on the utilized classification method, BCI systems can be categorized into two major groups, the traditional machine learning methods and deep learning approaches [13]. In the traditional machine learning approaches, the obtained features are fed into a traditional classifier like SVM [6, 9, 14], linear discriminant analysis (LDA) [5, 15], or k-nearest neighbor [16]. However, the deep learning approaches can take raw data as input and perform feature extraction and classification in a single network architecture [13]. The advantage of deep learning-based methods is that exclusive feature extraction is unnecessary since the network extracts relevant features to increase the system performance.

Both deep learning and traditional learning methods require a suitable EEG dataset for training the classifier. Due to the differences between various subjects, the classifier is trained based on an inter-subject protocol in most studies. On the other hand, recording EEG signals is a time-consuming and costly process, which might be frustrating for the subject under examination. Accordingly, the amount of training data is usually not very large. Therefore, reducing the data dimension is very important for training. This can be achieved through the selection of relevant channels. The channel selection may provide several additional benefits to the training process. For instance, it can eliminate noisy channels and reduce irrelevant data. It may also omit highly correlated channels with redundant information, thus, reducing the dimension of data and preventing the classifier's overfitting. It also leads to a reduction in time and hardware costs. Accordingly, a wide variety of feature and channel selection methods have been deployed for BCI systems.

Regarding evaluation criteria, channel selection methods are mainly divided into three categories: 1. filtering methods, 2. wrapper methods, and 3. hybrid methods [7]. Filtering techniques use evaluation criteria independent from the classification performance. Many channel selection methods are proposed in this category that come from classical data dimension reduction methods based on mathematical assumptions. For example, Common Spatial Pattern (CSP) is an effective spatial filtering method with many variants [17, 18]. All variants of these CSP-based spatiotemporal filtering such as SCSP [19], CSSP [17], FBCSP [20], STECS [5], TSGSP [6], and Riemannian-based approaches [21, 22] optimize a minimization problem of intra-class variance, or a maximization problem of inter-class covariance, or a combination of both. Then, channel or feature selection is applied either based on the weights of each channel (feature) in the transformed space [5, 18] or by calculating mutual information or other similar criteria [20].

Another instance can be found in [23], which introduces a metric based on prior information of the important frequency band for the task under the test to calculate each channel's importance. Das et al. [24] propose selecting channels with different mean and variance among classes under test and removing the others. Filtering methods improve the performance of BCI systems and overcome the dynamic nature of EEG signals. These improvements are caused by utilizing different spaces, multiscale time intervals, frequency ranges, and regularization. However, these methods have initial assumptions that limit their performance.

Wrapper techniques consider classification performance as the evaluation criterion and search among the feasible subset of channels to achieve a proper channel subset. Hence, they can create better classification results [7, 15]. However, these methods consume more time for system calibration. Hybrid methods try to benefit from both methods by combining them. For instance, a hybrid channel selection method is suggested in [25], which introduces a score called Dynamic Channel Relevance (DCR) based on the information theory metrics. The score is used iteratively in such a way to increase the relevance of selected channels while decreasing their redundancy. The DCR score is independent of the classification accuracy. Since this method is a combined one, the stopping condition for the channel selection process is defined based on the classification accuracy.

With regards to searching methods for a subset of channels, exact or approximate algorithms can be applied. Full search is an exact algorithm that guarantees to find the optimal subset. However, it is computationally costly, and the computation cost exponentially grows with the number of recorded channels, making it impractical. The sequential and metaheuristic methods are two categories of approximate algorithms, although their solution is sub-optimal [7, 26]. These methods require much less processing time and computational cost than the full search approaches. Sequential search methods start from an initial subset add or remove one component at a time to achieve better performance. They are typically the fastest approximate solutions. However, they suffer from premature convergence to local optimum, which is not desirable. Metaheuristic methods use randomness in the searching process to address the premature convergence problem, although it takes more time to converge. The evolutionary algorithms are a sub-category of meta-heuristic methods, such as variants of genetic algorithms [7, 15, 27], binary gravity search algorithm [28], and other algorithms [29, 30]. Evolutionary algorithms show superiority over mathematical channel selection methods.

In the previous wrapper methods, the performance of a traditional classifier is used as the evaluation criteria. Contrarily, in this work, we utilize a deep learning-based architecture for both feature extraction and classification. Hence, the evaluation criterion in channel selection is based on the performance of a deep-learning architecture. By this means, the whole process is entirely automated. Furthermore, we combine sequential and metaheuristic channel selection methods to merge their benefits and provide a faster and more efficient manner.

# 3. Proposed Method

A BCI system may involve several steps from signal acquisition to command generation, including preprocessing, feature extraction, channel selection, and classification [2]. Generally, these processes are applied in two phases: setup and daily application. The setup phase, which contains channel selection and system training, prepares the system for everyday applications. We have a trained system ready to apply the proposed method in daily use in the second phase. The test data in this work simulate the second phase. A block diagram of the proposed BCI system is presented in Figure 1. In both phases, the raw data is acquired and preprocessed in the first step to feed the next blocks. In the setup phase, after preprocessing, the channel selection step tries to find the best channel subset based on our proposed four-step channel selection method, which uses the accuracy of a deep neural network as the validation criterion. Then, the chosen channel subset is passed to the feature extraction. In the last step, classification is performed to estimate the class related to the input data.

In the application phase, the preprocessed test data feed the next block, where the selected channels from the setup phase are separated from the test data, and the modified test data is formed. Finally, the modified test data is transferred to the feature extraction and the classifier modules, the same network architecture used to validate channel subsets in the channel selection process. The weights of this network are calculated and saved in the setup phase during the channel selection process. Thus, we use the saved weights corresponding to the selected subset of channels for feature extraction and classification.

We can consider two different cases for choosing the desired number of channels during channel selection. The desired number of channels, $K$, can be predefined based on the hardware limitations. For example, if the user amplifier supports a certain number of channels, $K$ will be a fixed parameter. Another scenario is that there is no limitation for choosing the $K$, and any number of channels that lead to better performance can be used for daily use. As shown in Figure 1, in this study, $K$ is predefined.

In the following subsections, we initially explain how the stream of raw EEG data gets ready to pass to the next stages. Then, the proposed channel selection method is elaborated in subsections 3.2 to 3.5. Finally, feature extraction and classification are described in subsection 3.6.

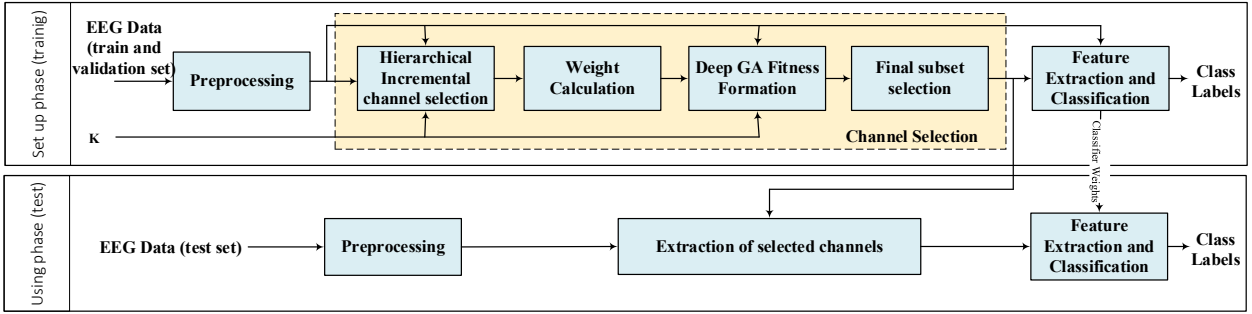

**Figure 1. Block diagram of the proposed method**

## 3.1. Preprocessing

We can clean the data to reduce the effects of noise and artifacts before other tasks are performed. To this end, in the first step after data acquisition, the data should be preprocessed. EEG signals can be easily affected with Electromyogram (EMG) and especially Electrooculogram (EOG) signals. Hence, firstly, trials that are extremely interfered with ocular and muscular signals are omitted. Depending on the used dataset, if it includes an EOG signal, various methods may be applied for this purpose or simply performed by an expert [31].

Then for each trial, the data is windowed in the time direction, as not all the time interval is informative for the classification task. Hence, considering the whole interval of the trial might decrease the classification accuracy. Hence, the interval is extracted from 0.5 seconds before the cue starts. Then the interval continues until the end of the motor imagery task [32]. The data is not filtered at the frequency domain in this step because it has been hardware filtered during the recording time. Then, data normalization is applied as the last step of data preparation for the following network. Since the conditions of data recording hardware are known, we have normalized the data between the maximum and minimum values corresponding to the possible range of dynamic changes in the recording phase. In other cases, the minimum and maximum values of training data can be used if there is no information about recording dynamic variations.

## 3.2. Hierarchical Incremental Channel Selection

Channel selection can increase the system's accuracy while reducing the complexity of the designed system [7]. For example, informative channels might be located near the corresponding area of the brain that oversees the function related to the task [7]. On the other hand, informative channels may not always be close, as close channels can be highly correlated. Consequently, choosing a proper criterion for channel selection is critical. Wrapper methods lead to higher accuracy since its evaluation criterion depends on the classifier's performance. This study uses a deep learning-based classifier as the evaluation criterion for channel selection.

One can use sequential or metaheuristic methods to generate subsets to find a channel subset with high performance. Sequential search methods are fast; however, they can get stuck in local extrema. The metaheuristic methods, such as genetic algorithms, check more subsets non-deterministically. Hence, they can find better subsets, although they require a longer processing time. Thus, bringing their benefits together may lead to a more efficient, faster, and more accurate solution. As shown in Figure 1, for a predefined value of $K$, the proposed channel selection method combines a sequential method with a metaheuristic algorithm in four steps. Here we will show how a hierarchical incremental channel selection algorithm (HICS) is applied in the first step as a sequential search method, which finds the best subset containing $K-1$ channels. In the second step, the output of the former step is used to calculate a weight vector (cf. subsection 3.3). As the third step, the proposed genetic algorithm called Deep GA Fitness Formation (DGAFF) plays the role of a metaheuristic method. DGAFF uses random weighted selection to generate the initial population. It uses the weight vector obtained from the previous step in initializing the first population and continues until the stopping criteria are reached. As the system starts from a better initial point, fewer generations are needed for converging to the final solution.

It is noteworthy that due to utilizing a network-based criterion for evaluation, the random nature of network training rooted in random initialization causes uncertainty and causes variations in the results of the fitness function for the same input. A subset history is considered and saves each subset of channels that appear during generations to address this problem. Weights of the network that is trained for each subset are also saved. So, if a subset is repeated in the reproduction of new generations, the previously saved corresponding weights are used to initialize the network weights rather than random initialization. The network is fine-tuned for a small number of epochs. As a result, the variation of the network accuracy is much smaller. Details of the algorithm can be found in subsection 3.4. The last thing to consider is that the proposed genetic algorithm may not converge to a single subset. Hence, the last step selects a proper subset among the last generation. In the following, we will explain the hierarchical search routine.

The hierarchical incremental search tries to find the most informative channel in the first step and then add other channels one by one until the stopping condition is satisfied. Stop criterion depends on the utilized scenario of channel selection. If a fixed number of channels is considered in the scenario, the algorithm continues to the determined number of channels. Otherwise, it continues until all the channels are aggregated in the final set. The best subset is saved for each number of channels while running the iterations. The final solution is the subset related to the highest accuracy among all the tested subsets. Therefore, the number of channels in the final subset can be between 1 and the total number of channels.

As illustrated in Figure 1, in the current work, a predefined fixed $K$ is considered as the number of channels. In the HICS, the selected channel number is one less than the predefined channel number ($K-1$). The pseudocode of the implemented algorithm is described in Algorithm 1. The preprocessed training and validation set with their labels and the predefined number of channels, $K$, are considered as the algorithm input. Variable C specifies the total number of channels in the dataset. In code lines 02 to 05, the primary subset $S_p$, final subset $S_f$ and temporary subset $S_t$ are initialized. The initial value for $S_p$ is the index of all the existing channels, while $S_f$ and $S_t$ are empty at the beginning.

Then, for each number of channels from $k=1$ to $k=K-1$, the following process is performed to find the best possible subset containing $k$ channels. Initially, a set for storing the evaluation result of each subset is initialized with the empty set, $Y \leftarrow \emptyset$ (line 07). Then, for all the existing channels in $S_p$, the concatenation of $S_f$ of the previous step, and the index of the current channel, $c_t$ is assigned to $S_t$ (line 10). The evaluation result of $S_t$ is calculated and concatenated with $Y$. So, in lines 10 to 12, each element of $S_p$ is concatenated to the resulting subset of the previous iteration in order to create $k$-channel subsets, while the corresponding evaluation value of each subset is stored in $Y$. In lines 14 and 15, the maximum value of the evaluation set, $Y$, is found. The channel corresponding to this value would be selected in the current iteration and concatenated to the $S_f$ to form the final set containing $k$ channels. This channel is also deleted from $S_p$ to avoid repeated channel subsets.

In this study, the evaluation of each channel subset is performed through training a network. The advantage of using deep neural networks is that, after performing the necessary preprocessing, such as filtering and normalization, the data can be fed into the network in raw form. Then the network will automatically perform feature extraction and classification simultaneously. Hence, there is no need for extracting handcrafted features, and the network will autonomously learn the informative features corresponding to the classification task.

In this work, EEGNet [32] is used to evaluate channel subsets. EEGNet is a convolutional architecture that is designed for the classification of EEG-based BCI systems. This network performs well in terms of accuracy, has a good generalizability, and is a lightweight network in terms of the number of parameters. Let's consider train and validation sets as $X_{train} \in \mathbb{R}^{N_{train} \times C \times T}$ and $X_{valid} \in \mathbb{R}^{N_{valid} \times C \times T}$, and their corresponding labels as $Y_{train} \in \mathbb{R}^{N_{train}}$ and $Y_{valid} \in \mathbb{R}^{N_{valid}}$, where $N_{train}$ is the number of training trials, $N_{valid}$ represents validation trials, $C$ is the total number of channels, and $T$ is the duration of each trial. For evaluation of a subset of channels containing $k$ channels, existing channels in that subset are extracted through all the trials. So, the modified data will be $\hat{X}_{train} \in \mathbb{R}^{N_{train} \times k \times T}$ and $\hat{X}_{valid} \in \mathbb{R}^{N_{valid} \times k \times T}$ (line 20). Then, a new model is created so that its input dimension for channels is set to $k$ instead of $C$. Afterward, the created model is trained using the modified training data, $\hat{X}_{train}$, and $Y_{train}$ (line 21). Finally, the accuracy of the model is obtained using the modified validation data, $\hat{X}_{valid}$, and $Y_{valid}$ and is considered as the evaluation criteria for the subset under test (lines 22 and 23).

Although sequential search methods are less complex than metaheuristic methods, they have deficiencies in their way of selection. If a channel is selected in one step, in the next steps, the existence of this channel is considered obvious. Hence, it will not be possible to evaluate other combinations of channels without the previously chosen ones.

If it was possible to check whether chosen channels in prior steps are proper for the current step, it could improve the system's performance. This is the motivation for combining this method with a metaheuristic approach.

---

**Algorithm 1:** Hierarchical Incremental Channel Selection (HICS)
**Inputs:**
  $K$: *The predefined number of channels*
  $((X_{train}, Y_{train}), (X_{valid}, Y_{valid}))$: *Training and validation sets with their corresponding labels*
**Output:**
  $S_{fHeuristic}$: *The output subset of channels resulted from HICS. It contains $K-1$ channels.*

**BEGIN**
01   $C \leftarrow$ length of $X_{train}$ in the channel dimension, representing the total number of channels
02   $S_p \leftarrow \{1, 2, \dots, C\}$
03   $S_{fHeuristic} \leftarrow \emptyset$
04   $S_t \leftarrow \emptyset$
05   $k \leftarrow 0$
06   **WHILE** ($k < K - 1$)
07     $Y \leftarrow \emptyset$
08     $k \leftarrow k + 1$
09     **FOR** each $c_t$ in $S_p$
10       $S_t \leftarrow S_{fHeuristic} \cup c_t$
11       $y \leftarrow$ **Evaluation_criterion**$(((X_{train}, Y_{train}), (X_{valid}, Y_{valid})), S_t)$
12       Y $\leftarrow$ **concatenate** $Y$ with $y$
13     **END**
14     $i_{max} \leftarrow$ **find** the index in which $Y = max(Y)$
15     $S_{fHeuristic} \leftarrow$ **concatenation** of the obtained channel, $S_p^{i_{max}}$, with "$S_{fHeuristic}$"
16     **delete** $i_{max}$<sup>th</sup> element of "$S_p$", $S_p^{i_{max}}$
17   **END**
18   Return($S_{fHeuristic}$)
**END**

---

**Procedure:** Evaluation_criterion
**Inputs:**
  $S_t$: *The subset of channels under test*
  $((X_{train}, Y_{train}), (X_{valid}, Y_{valid}))$ : *Training and validation sets with their corresponding labels*
**Output:**
  $y$: *The accuracy of EEGNet on the validation set corresponding to channels existing in subset $S_t$*

**Begin**
20   $(\hat{X}_{train}, \hat{X}_{valid}) \leftarrow$ **modify** $(X_{train}, X_{valid})$ according to $S_t$
21   $Model \leftarrow$ **train** EEGNet using $(\hat{X}_{train}, Y_{train})$
22   $\hat{Y}_{valid} \leftarrow$ **predict** $(\hat{X}_{valid}, Model)$
23   $y \leftarrow$ **accuracy**$(Y_{valid}, \hat{Y}_{valid})$
**END**

### 3.3. Weight Calculation

An intermediate block is required to combine HICS with the proposed genetic algorithm (DGAFF). In this module, a weight vector is built based on the selected $K-1$ channels in the HICS. This weight vector would be an input argument for DGAFF that uses weighted random selection for creating the initial population. The length of this vector is equal to the total number of channels existing in the dataset. Each element of this vector illustrates the probability of selection of its corresponding channel in the initial population of DGAFF. To allocate a higher probability to the selected channels in HICS for being selected in DGAFF, we assign a weight for these channels m times than the others. This is represented in the following equation.

$$w_i = \begin{cases} m, & \text{if } w_i \in S_{fHeuristic} \\ 1, & \text{otherwise} \end{cases} \quad (1)$$

In order to normalize these weights to use them as the probability of selection, we use the following equation.

$$\mathbf{w} = \frac{1}{\sum_{i=1}^{C} w_i} \begin{bmatrix} w_1 \\ w_2 \\ \vdots \\ w_C \end{bmatrix} \quad (2)$$

where, $\mathbf{w}$ is the weight vector with values in the range between 0 and 1. The higher the parameter $m$, choosing the selected channels from HICS is more probable. We have set $m = 2$ as a moderate value. This vector is passed to the next stage as the weight vector for weighted random selection of the initial population in the proposed genetic algorithm.

*3.4. Channel Selection based on DGAFF*

In this section, a genetic algorithm is proposed with a fixed number of desired channels which is set by user input. Algorithm 2 shows the pseudocode of this block. Each chromosome represents a subset of channels in a binary vector with a length of the total number of channels, $C$. Each element in a chromosome shows the presence or absence of the corresponding channel in the related subset. In the proposed genetic algorithm, the initial population contains $N_{fp}$ individuals, where each one is a vector with $K$ ones placed randomly in the vector (see Figure 2). The selection of the initial population is performed through a weighted random selection based on the output of the weight calculation block, the weight vector $\mathbf{w}$ (line 02 of Algorithm 2). In this process, the higher an element in the weight vector, the more likely it is to set its corresponding element to one in each individual vector of the initial population.

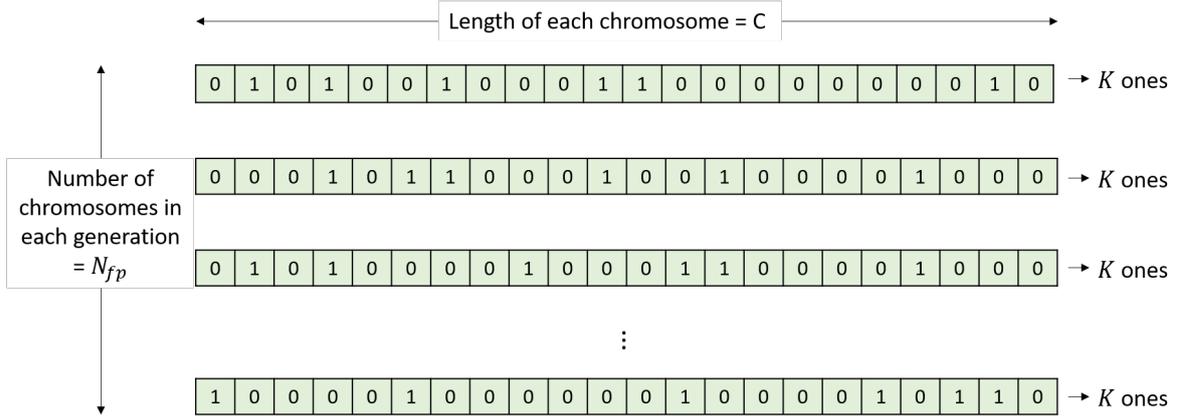

**Figure 2 Presentation of the chromosomes in an example generation. The number of ones in each chromosome is set to the predefined number of channels, $K$.**

Considering $P_k = \{s_k^j | j = 1, 2, \dots N_{fp}\}, k = 1, 2, \dots N_G$ as the population in the $k^{th}$ generation, $S_k^j$ is $j^{th}$ subset or $j^{th}$ individual of the population $P_k$, $N_{fp}$ represents the number of subsets in each population, and $N_G$ shows the maximum number of generations. Iterating through a loop for $N_G$ times (line 05), a new generation is iteratively created from the previous generation (lines 06-18). In each iteration of the algorithm, a certain number of pairs, $N_p$, are randomly selected from the current population members (line 09). Then each pair of subsets are combined with a probability $p_c$ (line 10), based on a two-point-cross-over function. Both of the resulting children are then mutated with a probability $P_m$ (line 11). The newly generated children are evaluated through the fitness function (line 12). The next generation is selected from the created offspring at the end of each generation, using the tournament method (line 16).

In our implementation, the fitness function is based on the validation accuracy of a network trained on the present channels in the related subset (individual). First, the number of ones in the generated children is checked to keep the number of channels fixed in all subsets, following the crossover and mutation steps (line 21). If they are not equal to the desired number of channels, the fitness value related to that individual is set to zero (line 27). Otherwise, validation accuracy of the corresponding trained EEGNet will be considered as the related subset's fitness value (lines 22-25).

**Algorithm 2:** Deep GA Fitness Formation (DGAFF)
**Inputs:**
    $K$: *The predefined number of channels*
    **w**: *The weight vector for weighted random selections of the initial population*
    $((X_{train}, Y_{train}), (X_{valid}, Y_{valid}))$: *Training and validation sets with their corresponding labels*
**Outputs:**
    $P$: *Final population*
    *Saved weights*: *The weights corresponding to the model trained for each subset in the final population*

**Parameters:**
$N_{fp}$: Number of individuals in the first population, 20
$N_G$: number of generations, 12
$N_p$: number of selected subset pairs in each generation, $round(7 \times N_{fp}/2)$
$P_c$: probability of cross over, 0.85
$P_m$: probability of mutation, 0.08

**BEGIN**
01    $C \leftarrow$ length of $X_{train}$ in the channel dimension, representing total number of channels
02    Weighted random generation of initial population "$P$" using $w$
03    Evaluate each individual of the first population with **Fitness_function**
04    $i \leftarrow 0$
05    **WHILE** ($i < N_G$)
06        $P_{new} = \emptyset$
07        $Y_{new} = \emptyset$
08        **FOR** $j$ in **range**($N_p$)
09            random selection of $P_1$ and $P_2$ form $P$
10            **Cross over** $P_1$ and $P_2$ to get $NP_1$ and $NP_2$
11            **Mutation** of $NP_1$ and $NP_2$
12            **Evaluation** of $NP_1$ and $NP_2$
13            **Concatenation** of "$P_{new}$" with $NP_1$ and $NP_2$
14            **Concatenation** of "$Y_{new}$" with the corresponding fitness values to $NP_1$ and $NP_2$
15        **END**
16        selection of "$N_{fp}$" number of "$P_{new}$" based on their fitness value in $Y_{new}$
17        set "$P_{new}$" as the first population, "$P$"
18        $i \leftarrow i + 1$
19    **END**
20    Return($P$, $Saved\ weights$)
**END**

**Procedure:** Fitness_function
**Inputs:**
    $P_j$: *The subset of channels under test*
    $K$: *The predefined number of channels*
    $((X_{train}, Y_{train}), (X_{valid} \in, Y_{valid}))$: *Training and validation sets with their corresponding labels*
**Output:**
    $y$: *The accuracy of EEGNet on the validation set corresponding to channels existing in subset $P_j$*

**Begin**
21    **IF** summation of elements in $P_j = K$
22        $(\hat{X}_{train}, \hat{X}_{valid}) \leftarrow$ **modify** ($X_{train}, X_{valid}$) according to $S_t$
23        $Model \leftarrow$ **train** EEGNet using $(\hat{X}_{train}, Y_{train})$
24        $\hat{Y}_{valid} \leftarrow$ **predict** ($\hat{X}_{valid}, Model$)
25        $y \leftarrow$ **accuracy**($Y_{valid}, \hat{Y}_{valid}$)
26    **ELSE**
27        $y \leftarrow 0$
28    **END**
**END**

*3.5. Subset Selection*

In this study, he convergence of the proposed genetic algorithm has been a challenge for two reasons. First, it may take numerous iterations, and thus the training can be a time-consuming process. Second, the fitness function is a network. Due to random initialization of its weights, the accuracy varies each time the network is trained, even if the channel subset is the same. Hence, we are dealing with a fitness function changing in each execution. This is another challenge in the convergence of the genetic algorithm. To deal with these problems, we have limited the algorithm to a small number of generations ($N_G$). However, after the number of generations reaches $N_G$, we need to select a proper subset among the population of the last generation. For this goal, two criteria are utilized:
1. Fitness values correspond to each individual because the fitness value is the accuracy of the network given the subset of channels that individuals indicate. Therefore, the subsets of channels with higher fitness values are more likely to cause higher accuracy in the classification step.
2. The number of each subset repetitions in the last generation, as an individual's repetition in a generation implies that the algorithm is going to converge to that individual in the latest generations.

For combining these two criteria, the unique subsets are drawn out of the last generation. We assume there are $N_U$ unique individuals among $N_{fp}$ individuals of the last generation, $P_{N_G}$, that each one is represented by $S^i$. Then for each $S^i$, two values are itemized, including the number of repetitions of the subset and its corresponding accuracy. Therefore,

$$N_{fp} = \sum_{i=1}^{N_U} r^i \qquad (3)$$

where, for each $S^i$, $r^i$ is its repetition in $P_{N_G}$. We can normalize each $r^i$ to get

$$\hat{r}^i = \frac{r^i}{N_{fp}} \qquad (4)$$

where, $\hat{r}^i$ is the normalized $r^i$. Let us assume that $y^{S^i}$ is the fitness value of $S^i$. Since each $y^{S^i}$ represents the accuracy of an individual, it is in the range of [0, 1]. Therefore, the two values, $y^{S^i}$ and $\hat{r}^i$ are mapped in the range of [0, 1] which makes the comparison of their effect simpler. Then, we can combine $y^{S^i}$ and $\hat{r}^i$ using the following equation

$$\gamma \times \hat{r}^i + (1 - \gamma) \times y^{S^i} \qquad (5)$$

In this equation, the $\gamma$ parameter determines the ratio of repetition and accuracy and, it can be chosen from the range of [0, 1] by the system designer. Less value of $\gamma$ means that the value of accuracy has higher importance in comparison with repetition of the subsets and vice versa. The selected subset of channels among the last generation, $S^f$, is chosen as the following:

$$S^f : f = \arg\max_i \left( \gamma \times \hat{r}^i + (1 - \gamma) \times y^{S^i} \right), \qquad i = 1, \dots, N_U \qquad (6)$$

*3.6. Feature Extraction and Classification*

Following the previous steps, including the data recording, preprocessing, and channel selection, the selected channels are extracted and utilized for the classification task. EEGNet is utilized in the last step as the feature extractor and classifier, as illustrated in the final block of Figure 1. The channels that do not exist in the selected subset are removed from all the train and test sets samples. Then, the modified data is fed into the network. Finally, the network is trained using the train set, and the accuracy of the test set is reported as the performance of the resulting subset.

## 4. Experimental Results and Discussions

For the implementation of the proposed methods, Python language is used. EEGNet[1] is implemented based on Keras API on top of Tensorflow. In the following sections, we initially introduce the dataset used to evaluate our method. In subsection 3.2, the hyper-parameters used to implement the network and channel selection methods are reported. Also, the effect of changing the $\gamma$ parameter in selecting a final subset of channels among candidate subsets in the last generation of DGAFF is studied. In subsection 3.3, the relation between selected channels and brain parts is investigated. Finally, in subsection 3.4, the comparative results are presented. Each part of the proposed method is implemented individually. The results are compared to show the superiority of their combination. Then, the proposed combined method is compared with other works in this field.

---
[1] https://github.com/vlawhern/arl-eegmodels

## 4.1. Dataset Description

The EEG data used in this work is dataset IIa of BCI Competition IV [31]. This dataset is a 4-classes motor-imagery dataset. During the signal acquisition step, the arrow-like cues pointing to cardinal directions are used to indicate the imagination of movement of the left hand, right hand, feet, and tongue. The data is acquired from nine healthy subjects through 22 electrodes that are located according to the 10-20 system. The data includes two sessions recorded on different days for each subject, and each session contains 288 trials. The first session is used as the training data, and the second one is considered as the test data. In this scenario, a fixation cross is initially shown on the screen for recording the data for each trial. It continues for 2 seconds, and a short beep sounds at the interval's beginning. Then an arrow pointer is shown on the screen that tells the subject which movement he should imagine. The cue stayed on the screen for about 1.5 seconds. Then the subject should imagine the desired action till the sixth second of the trial time. After that, a short break is considered before the subsequent trial begins.

The data is sampled with 250 Hz and a band-pass filter between 0.5 Hz and 100 Hz, and a 50 Hz notch filter is applied to it. The sensitivity of the amplifier is set to 100 μV during data recording. Thus, the data is normalized between the minimum and maximum values in the preprocessing step. Based on the hardware settings, the range of data is considered ±100. Besides, each trial has a flag annotated by an expert, which indicates whether the trial is affected by intense noise and artifacts or not. This flag is used to eliminate damaged trials towards achieving higher SNR. To prevent leakage of test data in the channel selection and classification process, we separate 20 % of the first day's data as an evaluation set. The accuracy of this part is used in the training process for channel selection. The validation set is the same in each training process.

## 4.2. Parameter Analysis

The hyper-parameters utilized for the network and the genetic algorithm are presented in this section. In the EEGNet architecture, numbers of temporal and spatial filters, $F_1$ and $F_2$ are set to 8 and 16, respectively. Also, the depth multiplier, $D$, is set to 2, the same as their default values in [33]. The probability of the dropout layer has been changed from 0.25 to 0.1. In DGAFF, the length of each chromosome is 22, which is equal to the total number of channels in the dataset. Furthermore, the probability of crossover is set to 0.85, and the probability of mutation is set to 0.08. The number of individuals in the initial population is set to 12. In each generation, 42 offspring are generated, from which only 12 will survive and be passed to the next generation.

Another hyper-parameter that should be discussed is the $\gamma$ parameter. As mentioned earlier, $\gamma$ is a hyper-parameter that tunes the ratio of model accuracy and repetition of each subset in the final generation for selecting the proper subsets. If $\gamma$ is set to zero, only the accuracy is considered for optimization. If $\gamma$ equals one, the repetition of subsets is the only criterion for the selection of the proper subset from the last generation. For the proposed combined method to extract six channels, different values from 0 to 1 with intervals of 0.2 are tested as $\gamma$, and the best results in average on the validation set are obtained when $\gamma$ is 0.3. These results are reported for all of the subjects in Table 1. The most accurate subsets are often repeated more frequently in the final generation. Hence, these two criteria correlate together, and different values for $\gamma$ makes little changes in their combination.

Table 1. Investigating the effect of parameter $\gamma$ in the performance of the proposed method.

| Sub \ $\gamma$ | 1 | 2 | 3 | 4 | 5 | 6 | 7 | 8 | 9 | Avg |
|---|---|---|---|---|---|---|---|---|---|---|
| 0 | 87.80 | 62.11 | 94.89 | 71.03 | 93.19 | 74.26 | 86.57 | 94.63 | 88.15 | 83.63 |
| 0.1 | 86.81 | 61.97 | 96.35 | 71.72 | 92.59 | 74.26 | 87.14 | 98.21 | 86.62 | 83.96 |
| 0.3 | 87.38 | 61.69 | 97.23 | 71.21 | 92.15 | 77.41 | 87.14 | 97.91 | 87.54 | **84.41** |
| 0.5 | 87.23 | 61.97 | 96.64 | 70.86 | 92.74 | 76.67 | 87.43 | 98.36 | 87.08 | 84.33 |
| 0.7 | 84.68 | 61.97 | 94.31 | 67.24 | 85.63 | 75.37 | 89.86 | 97.61 | 82.08 | 82.08 |
| 0.9 | 83.4 | 63.8 | 91.39 | 66.38 | 84.89 | 75.56 | 89.86 | 97.31 | 81.57 | 83.4 |
| 1 | 87.23 | 61.97 | 97.08 | 70.17 | 93.19 | 76.85 | 87.43 | 98.36 | 86.92 | 84.36 |

*4.3. Comparative Results*

The proposed method in our work combines a sequential search method with a genetic algorithm that tries to take advantage of the benefits of both methods. The genetic algorithm (GA) is a metaheuristic method with a higher search scope. GA is combined with hierarchical incremental search, which is much faster. Table 2 demonstrates the total effect of channel selection in terms of accuracy. Also shown in Table 2 is the influence of each constitute module on the performance of the proposed method. In the first row of Table 2, EEGNet is trained using all 22 channels existing in the dataset for each subject. The accuracy in each case is reported. In the next three rows of Table 2, the results of the three different scenarios are compared. These scenarios are implemented to choose six channels. The minimum number of channels selected in [28] is six. Only HICS is used for channel selection in the first scenario, and other blocks are bypassed. In the second scenario, only DGAFF is applied for channel selection. In this case, the weight vector cannot be calculated, as the HICS block is not used. So, the initial population is selected randomly. The third scenario is the proposed combined channel selection method and applies all the blocks. Table 2 shows that the accuracies of DGAFF and the proposed method are higher than the EEGNet model trained for all the channels. The HICS has a slightly lower accuracy while reducing the used channels from 22 to 6. Furthermore, the superiority of the proposed method is shown compared to each constitute module.

**Table 2. Comparison of proposed methods, HICS, DGAFF, and combined, in terms of classification accuracy**

| subject | 1 | 2 | 3 | 4 | 5 | 6 | 7 | 8 | 9 | Average |
|---|---|---|---|---|---|---|---|---|---|---|
| EEGNet [33] | 80.85 | 67.89 | 82.04 | 65.51 | 89.33 | 71.66 | 89.14 | 91.42 | 82.77 | 80.07 |
| HICS | 80.14 | 59.43 | 73.96 | 89.33 | 89.33 | 58.7 | 84.85 | 91.04 | 79.08 | 78.43 |
| DGAFF | 87.80 | 64.51 | 88.18 | 72.93 | 77.04 | 82.59 | 88 | 90.15 | 87.08 | 82.03 |
| Proposed combined method | 87.38 | 61.69 | 97.23 | 71.21 | 92.15 | 77.41 | 87.14 | 97.91 | 87.54 | 84.41 |

Our proposed method is compared with the channel selection methods used in [28] in this subsection. In Table 3 and Table 4, our proposed method is executed with the number of channels used in [28] for a fair comparison. In Table 3, our proposed method is compared with the metaheuristic channel selection method in [28], which is applied based on the binary gravitational search algorithm. This method is a wrapper method in which the evaluation of each subset of channels is performed based on an SVM classifier. We see that our proposed method has higher accuracy for all the subjects in the dataset except subjects 2 and 4. Also, our proposed method has performed better by an average of 5.46% compared to the method presented in [28].

**Table 3. Comparison of classification accuracy of the proposed method with metaheuristic method proposed in [28]**

| Subject | Accuracy of the proposed method | Accuracy of the method in [28] using IBGSA | K |
|---|---|---|---|
| 1 | 87.80 | 76.66 | 9 |
| 2 | 62.11 | 76.66 | 6 |
| 3 | 94.89 | 73.33 | 7 |
| 4 | 71.03 | 73.33 | 8 |
| 5 | 93.19 | 80 | 11 |
| 6 | 74.26 | 73.33 | 12 |
| 7 | 86.57 | 76.66 | 15 |
| 8 | 94.63 | 80 | 11 |
| Average | 83.63 | 76.24 | 9.87 |

In [28], the proposed metaheuristic method, IBGSA, is compared with dimension reduction using the PCA method. In the second method presented in [28], the dimensionally reduced data is then classified using SVM and DADA classifier, shown in columns 2 and 3 of Table 4. Since the number of components used in PCA was not the same as the number of IBGSA channels, our proposed method is implemented with the number of components resulting from data reduction by PCA. As shown in Table 4, for an equal number of channels, the proposed method has performed better for all subjects except for subject 2. For the second subject, PCA dimension reduction combined with SVM classification improves performance. It is worth noting that in [28], the data is preprocessed in several stages, and artifacts are decreased. For subject 2, the poor classification accuracy shows that this subject's data is very noisy. Probably, the reason for improving the performance of the methods presented in [28] compared to our proposed method is the preprocessing stages, as the only preprocessing used in our approach is normalization. Consequently, it is expected that in the proposed method, if the preprocessing stage is strengthened, the system performance can improve.

**Table 4 Comparison of classification accuracy of the proposed method with PCA-based methods proposed in [28]**

| Subject | Accuracy of the proposed method | Accuracy of the method in [28] using PCA with DQDA classifier | Accuracy of the method in [28] using PCA with SVM classifier | K |
|---|---|---|---|---|
| 1 | 83.55 | 53.33 | 76.66 | 14 |
| 2 | 63.24 | 53.33 | 63.33 | 18 |
| 3 | 94.89 | 60 | 66.66 | 7 |
| 4 | 75.86 | 60 | 53.33 | 7 |
| 5 | 88.74 | 63.33 | 56.66 | 8 |
| 6 | 77.78 | 56.66 | 66.66 | 20 |
| 7 | 89.86 | 53.33 | 40 | 10 |
| 8 | 97.61 | 53.33 | 56.66 | 16 |
| Average | 83.94 | 56.66 | 60 | 12.5 |

The methods presented in [23], [24], and [19] may be considered as filtering methods. The proposed method in [23] assigns a weight to each channel. These weights are defined based on the power of specific frequency bands and certain time intervals. The weights of the channels are repeatedly updated according to the distance from a reference area. After a particular number of iterations, channels that have higher weights remain. In the proposed method of [24], which is called ECSP, the channel selection criterion is determined by the difference between the mean and the standard deviation of the two classes existing in the train data. Also, in [19], two types of norms are used to regularize the weights of CSP transform. Hence, the sparsity of the weights increases in this space.

Our proposed method is compared against these three studies, as demonstrated in Table 5. Then, the results of our proposed method for selecting six channels are presented. In methods of [23], [24], and [19], the accuracy for each subject is shown. The last row shows the average number of channels over nine subjects. It can be seen that our proposed method, on average, has performed better than all subjects existing in the dataset. Also, the number of channels we have chosen in our proposed method is less than the other methods.

**Table 5 Comparison of classification accuracy of the proposed method with the methods presented in [23], [24], and [19]**

| subject | Accuracy of the proposed combined method | Accuracy of the proposed method in [23] | Accuracy of ECSP [24] | Accuracy of SCSP1 [19] | Accuracy of SCSP2 [19] |
|---|---|---|---|---|---|
| 1 | 87.38 | 89.65 | 83.36 | **91.66** | **91.66** |
| 2 | 61.69 | 62.96 | **71.83** | 67.36 | 60.41 |
| 3 | 97.23 | 95.81 | **98.54** | 97.91 | 97.14 |
| 4 | 71.21 | **75.03** | 74.13 | 72.22 | 70.83 |
| 5 | **92.15** | 87.8 | 71.11 | 65.27 | 63.19 |
| 6 | **77.41** | 66.29 | 73.14 | 66.67 | 61.11 |
| 7 | 87.14 | **89.06** | 83.57 | 84.72 | 78.47 |
| 8 | **97.91** | 94.53 | 96.26 | 97.22 | 95.13 |
| 9 | 87.54 | 69.76 | **94.61** | 91.66 | 93.75 |
| Average | **84.41** | 81.21 | 82.95 | 81.63 | 79.07 |
| K | 6 | 10 | 8.11 | 13.22 | 8.55 |

### 4.4. Channel Location Analysis

In a BCI system, the placement location of electrodes is very important as it affects the performance of the system. Besides, each part of the brain is more involved in specific functions and tasks, while there are variations between different subjects and a subject in different time intervals [3, 8, 15]. The dataset used in this work has 22 channels which are mainly located in the center of the brain, where is mentioned to be involved in motor imagery tasks [34]. The placement of electrodes and their location based on standard segmentation of the brain is shown in Figure 3.

Since channel selection is applied for each subject separately, the final subset for each subject is different. Figure 4 presents a histogram of the existence of each channel in the final output of all subjects that shows which channels are more general between different subjects. It can be observed that channels 18, 5, 1, 14, and 19 are selected more frequently in the final output of different subjects. Channel 18, 19, and 14 are located around the supramarginal gyrus, and channel 5 is on the premotor cortex. Channel 1 is placed on the frontal lobe, which involves high-order sensorimotor actions in healthy participants [34, 35].

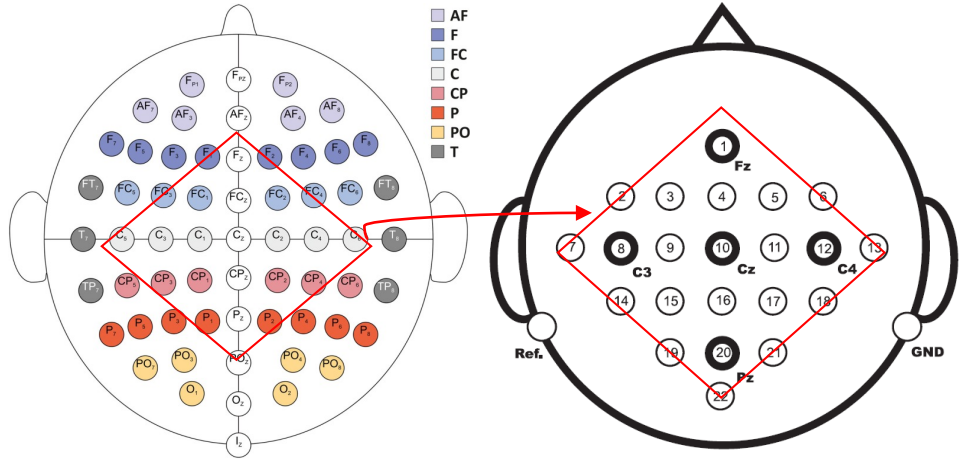

**Figure 3. Location of the used channels in the dataset [31] based on international 10-20 system [36].**

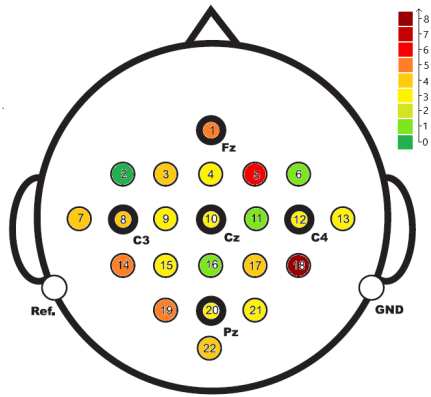

**Figure 4. Topographical mapping of selected channels. Frequently selected channels in different subjects have higher color intensity.**

# 5. Conclusion

Channel selection is an essential task in BCI systems that can increase classification performance while reducing the complexity of the model. Furthermore, it may help alleviate the inter-subject variations. In this work, a wrapper method including a combined strategy of incremental search algorithm and genetic algorithm is used to aggregate the benefits of both methods. First, an incremental search algorithm helps find channel subsets with fewer training steps than the genetic algorithm. Then, DGAFF is initialized with this subset and finds the final subset of channels with fewer generations than random initialized DGAFF and a higher scope than the incremental search alone. The proposed method enhances classification performance by 5.69% on average compared against the other channel selection methods on BCI-Competition IV dataset IIA.

A simple network called EEGNet is used with fewer (2.6k) trainable parameters in the classification step. In the future, better networks can be utilized combined with proposed channel selection methods that increase classification accuracies in the channel selection phase.